\documentclass[letterpaper, 10 pt, conference]{ieeeconf}  

\IEEEoverridecommandlockouts                              

\usepackage{amsmath} 
\usepackage{amssymb}  
\usepackage{multirow}
\usepackage{graphicx}
\usepackage{float} 
\usepackage{xcolor}
\newcommand{\etal}{\textit{et al.}}
\usepackage{stfloats}
\usepackage[caption=false]{subfig}
\usepackage{siunitx}

\title{\LARGE \bf
Shape Formation for the Cooperative Transportation of Arbitrary Objects Using Multi-Agent Reinforcement Learning
}

\author{Mohamed Sayed$^{1}$, Wolfram Burgard$^{1}$ and Tanja Katharina Kaiser$^{1}$
\thanks{$^{1}$All authors are with the Department of Computer Science and Artificial Intelligence, University of Technology Nuremberg, Germany. {\tt\small mohamed.sayed@utn.de}}
}%

\begin{document}

\maketitle
\thispagestyle{empty}
\pagestyle{empty}

\begin{abstract}
Cooperative object transportation is essential in numerous domains, including industrial to domestic services.
A popular transportation strategy is to carry objects on top of multi-robot systems. The corresponding task is typically solved by decomposing it into 
three interconnected subproblems: formation control, cooperative navigation, and collision avoidance.
A particular challenge posed by real-world objects is their potentially arbitrary shape and non-uniform mass distribution, necessitating robot formations that securely support the object.
In this work, we address the challenge of pattern formation control for transporting such real-world objects by proposing a novel multi-agent reinforcement learning approach.
Our approach enables a multi-robot system to autonomously position itself underneath an object to support its weight while avoiding obstacles during the formation process.
Our evaluations with diverse environments and varying numbers of robots show that our approach leads to policies that reliably produce balanced formations and generalize to cluttered scenes and objects with complex geometry and non-uniform mass distribution.
\end{abstract}

\section{INTRODUCTION}

Cooperative object transport is a central problem in the space of multi-robot systems (MRSs) with relevant applications in various domains, including logistics and warehouse automation. 
Existing strategies include pushing, grasping, and caging~\cite{tuci2018}.
A different strategy is transporting an object on top of robots. 
A prominent example of this strategy are systems like Amazon’s Kiva robots~\cite{wurman2008kiva}.  
In such systems, each robot transports a single, manageable unit (e.g., a shelf) by positioning itself underneath and lifting it. 
While effective, this approach relies on a structured interface between the robot and the object: shelving units and robots are designed to match in terms of form factor, lifting mechanism, and physical capabilities.
Another approach is placing the object on top of a flexible sheet or platform that is jointly supported by several robots, which requires explicit modeling of tension and compliance effects~\cite{hardy2021elastic}. 
While effective in controlled settings, these methods typically impose strong assumptions on the object’s geometry, mass distribution, or interface constraints. 
Moreover, the existing strategies primarily focus on navigation during transport, rather than on the prior challenge of forming a stable load-bearing formation beforehand. 
However, everyday objects often have irregular shapes and non-uniform mass distributions.
For the transportation of such objects, we cannot rely on specifically designed interfaces and require formation control approaches that generalize to arbitrary objects.  

\begin{figure}[t]
    \centering
    \subfloat[]{
        \includegraphics[width=0.31\linewidth]{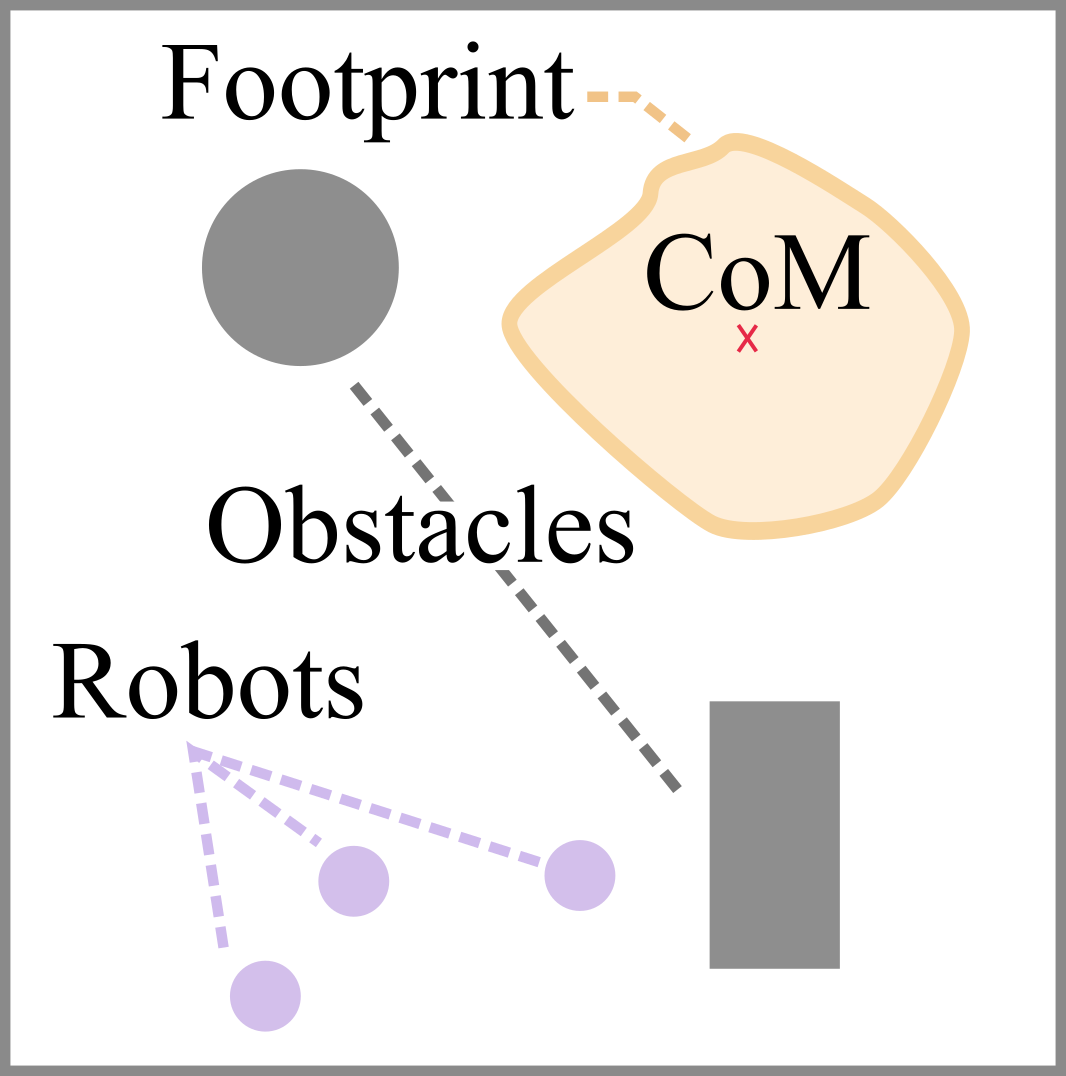}
        \label{fig:01a}
   }
    \subfloat[]{
        \includegraphics[width=0.31\linewidth]{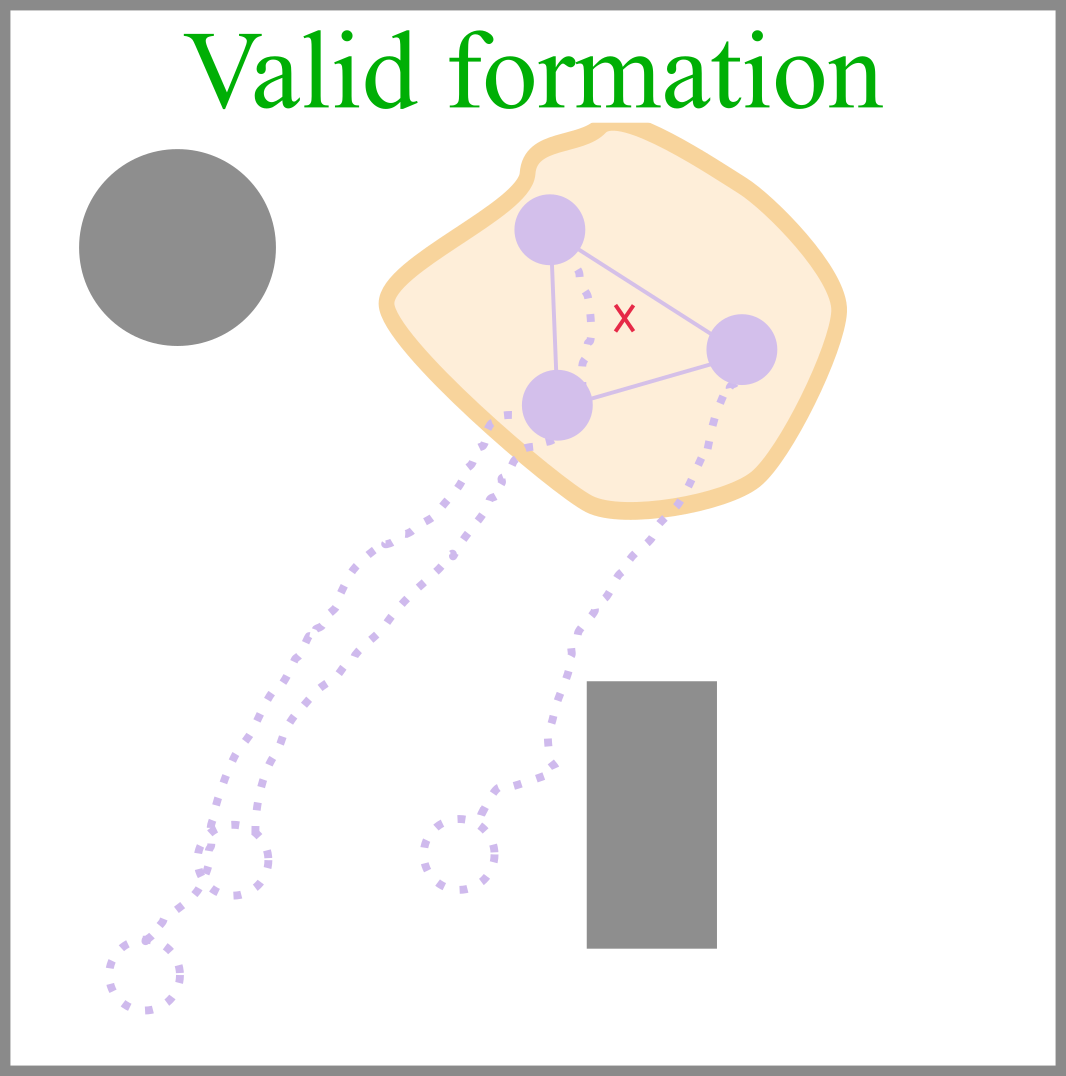}
        \label{fig:01c}
    }
    \subfloat[]{
        \includegraphics[width=0.31\linewidth]{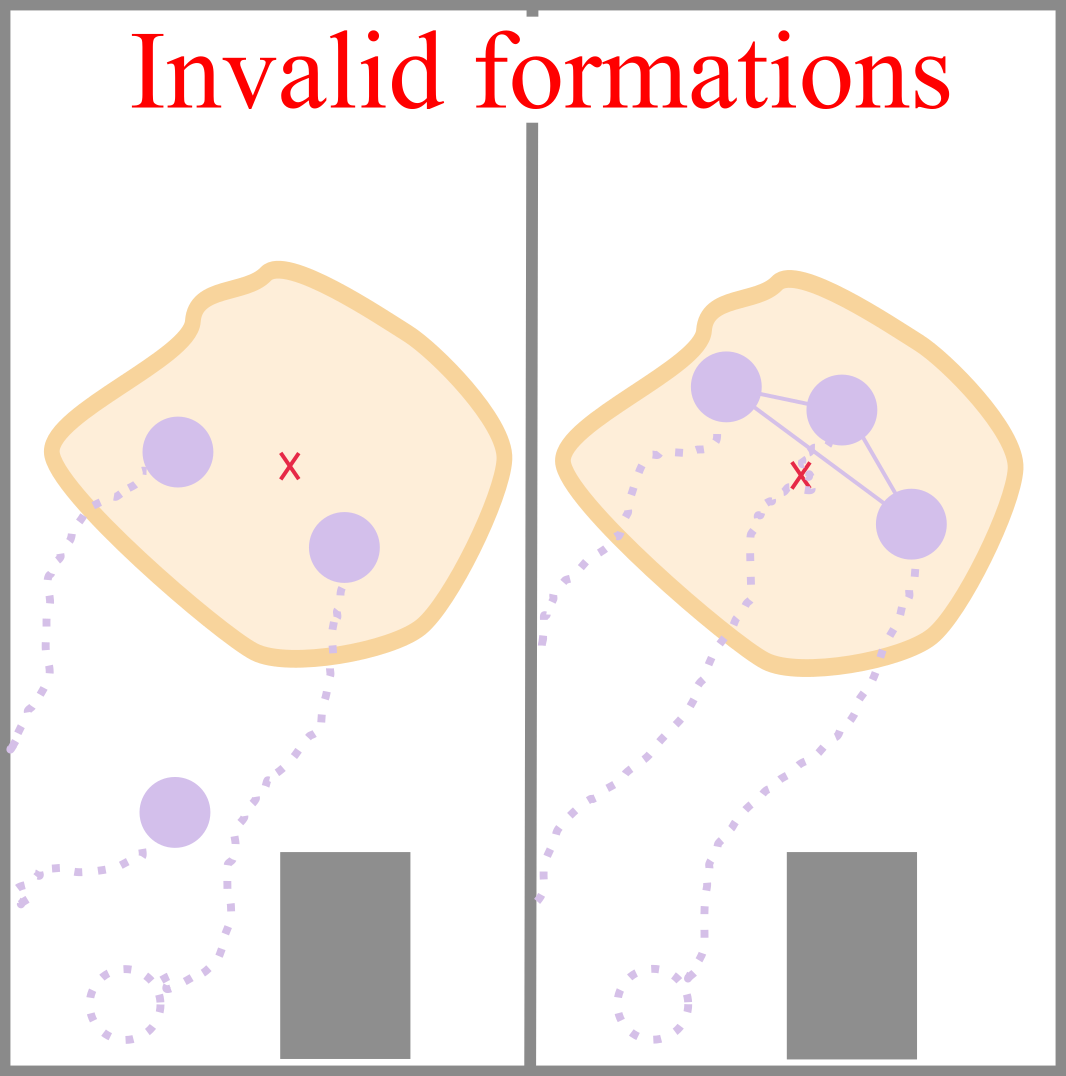}
        \label{fig:01b}
    }    
\caption{Overview of formation task for object carrying with (a)~the scene description, as well as examples of (b)~a valid formation and (c)~invalid formations.}
    \label{fig:overview}
\end{figure}

In this work, we focus on the initial phase of transport: forming a stable load-bearing formation beneath a passive rigid object of arbitrary shape and mass distribution (see Fig.~\ref{fig:overview}). 
The robots start from random initial positions and must self-organize within the object's footprint to achieve an even load distribution. 
A balanced initial formation is essential for safe and reliable transport afterward.
To the best of our knowledge, we are the first to explicitly address the pattern formation problem for transporting objects with non-uniform mass distribution.
 
We propose a multi-agent reinforcement learning (MARL) approach based on the centralized training decentralized execution (CTDE) paradigm in the physics-based simulation environment VMAS~\cite{bettini2024vmas}.
Robots learn decentralized policies that enable them to form load-aware support structures within an object's footprint while avoiding collisions with each other and obstacles during the pattern formation process. 
Our experiments demonstrate that the learned policies generalize across cluttered environments, different object geometries and mass distributions, and varying numbers of robots.

In summary, we make the following contributions: 
\begin{enumerate}
    \item We introduce a cooperative object transport scenario focused on load-bearing pattern formation under rigid objects with arbitrary shape and mass distribution.
    \item We propose a simplified abstraction of the load-sharing problem using local load estimates and limited object footprint information.
    \item We train decentralized MARL policies that enable autonomous pattern formation, balanced load distribution, and collision avoidance.
    \item We demonstrate scalability and generalization across environments and object complexities. 
\end{enumerate}

\section{RELATED WORK}

\subsection{Cooperative Object Transport}

The task of transporting objects on top of a multi-robot system has been intensively studied in the past. It requires formation control, cooperative navigation, including formation marching, and reliable obstacle avoidance. 

Several previous works studying this transportation task rely on  optimization methods~\cite{Koung2021HQP}, finite state machines~\cite{jurt2022collective}, or genetic programming and decentralized negotiation strategies~\cite{Herranz2022Negotiation}. 
The transported object may either be firmly attached to the robots~\cite{Herranz2022Negotiation,Koung2021HQP}, or loosely placed on top of the multi-robot system~\cite{jurt2022collective}.   
Some works assume a uniform mass distribution~\cite{jurt2022collective}, which  simplifies the formation control task by requiring robots to be approximately uniformly distributed. 
Others predefine target formations via robot-object attachment points~\cite{Herranz2022Negotiation} or inter-robot distances~\cite{Koung2021HQP}, thereby enabling the transport of objects with varying mass distributions.
Although the robot formation is crucial for transporting objects on top of several robots, these works do not address how the target formation itself is determined. 
To the best of our knowledge, we are the first to explicitly focus on formation control under different mass distributions in the context of cooperative object transport. 

Pattern formation studies investigate how robots can arrange themselves into specific shapes, either by outlining~\cite{Wang2019} or by filling them~\cite{Rubenstein2014,Liu2024Self-Healing,Bi2018SOSF}. 
Existing approaches include algorithmic methods~\cite{Rubenstein2014,wang2020taskswapping}, potential field techniques~\cite{Bi2018SOSF}, Large Language Models~\cite{venkatesh2024zerocap}, and reinforcement learning~\cite{Wang2019}. 
However, these methods focus primarily on geometric shapes, rather than on mass distribution.
Liu et al.~\cite{Liu2024Self-Healing} incorporate the center of mass (CoM) through an image moments-based approach.
Nevertheless, different shapes can have the same CoM, and robots may end up outside the target shape in their approach.

In contrast to these previous methods, our approach aims at achieving a uniform load distribution within the object footprint with all robots of the multi-robot system to ensure safe and stable transport. 

\subsection{Multi-Agent Reinforcement Learning}

\begin{figure}[t]
    \centering
    \includegraphics[width=0.9\linewidth]{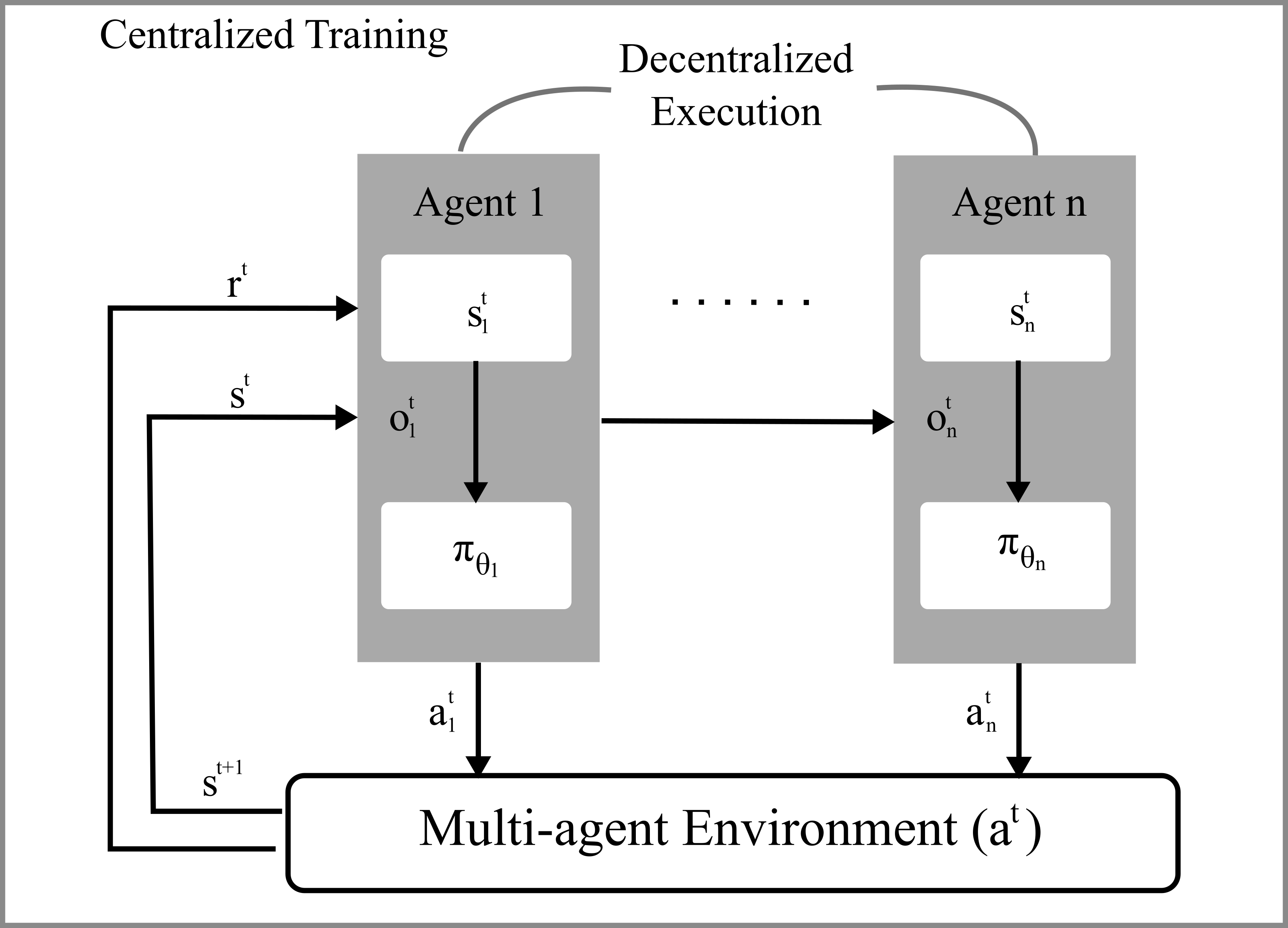}
    \caption{Centralized Training Decentralized Execution Paradigm}
    \label{fig:algorithm}
\end{figure}

Learning-based approaches to robot control enable generalization across environments and address hard-to-engineer problems. 
In reinforcement learning (RL), robots train policies through trial and error, aiming to maximize a cumulative reward.
Multi-Agent Reinforcement Learning (MARL)~\cite{marl-book} extends this approach to multi-agent systems, introducing additional challenges such as non-stationarity, scalability, and credit assignment.
The centralized training decentralized execution (CTDE) paradigm (see Fig.~\ref{fig:algorithm}) addresses these challenges by improving learning efficiency while maintaining scalability~\cite{du2021survey}.
During training, agents have access to shared or global information, whereas during execution, they solely rely on local observations.

MARL has been applied to multi-robot systems in tasks such as order picking in warehouses~\cite{krnjaic2023scalable}, autonomous driving~\cite{zhou2020smarts}, and navigation~\cite{Juntong2019}.
Several works have also applied MARL to the pattern formation problem. 
Iiama \etal~\cite{Iima2015MARL} train agents in a 2D grid world to form patterns as quickly as possible, while 
Wang \etal~\cite{Wang2019} propose a CTDE-approach to train general-purpose pattern formation policies that can adapt to new formations with only minimal additional training. 
However, these methods rely on predefined target poses and do not consider object mass distributions.
In contrast, we do not specify fixed robot positions in advance.

\section{METHODS}

\subsection{Task Description}

In an environment with up to \(n_O\)~obstacles, a decentralized multi-robot system of \(N\)~robots must arrange itself into a formation in order to carry an object of arbitrary shape and mass distribution on top of it. 
The object and \(n_O\)~obstacles are at random poses in the environment.
We assume the object to be hovering over the robots, similar to being on stilts. 
Thus, robots cannot collide with the object.
The footprint of the object is given by a 2D polygon \(\mathcal{P} \subset R^2\). 
We use three types of objects: simple ones with uniform density, more complex ones with non-uniform density, featuring both heavy and light regions, and complex objects with cavities and non-uniform density.
The object's center of mass \(CoM = (x_c, y_c)\) is computed geometrically, accounting for the varying densities within the object, and is relevant with respect to the load distribution among robots.

\begin{figure}[t]
    \centering
    \subfloat[Local Observation]{
        \includegraphics[width=0.49\linewidth]{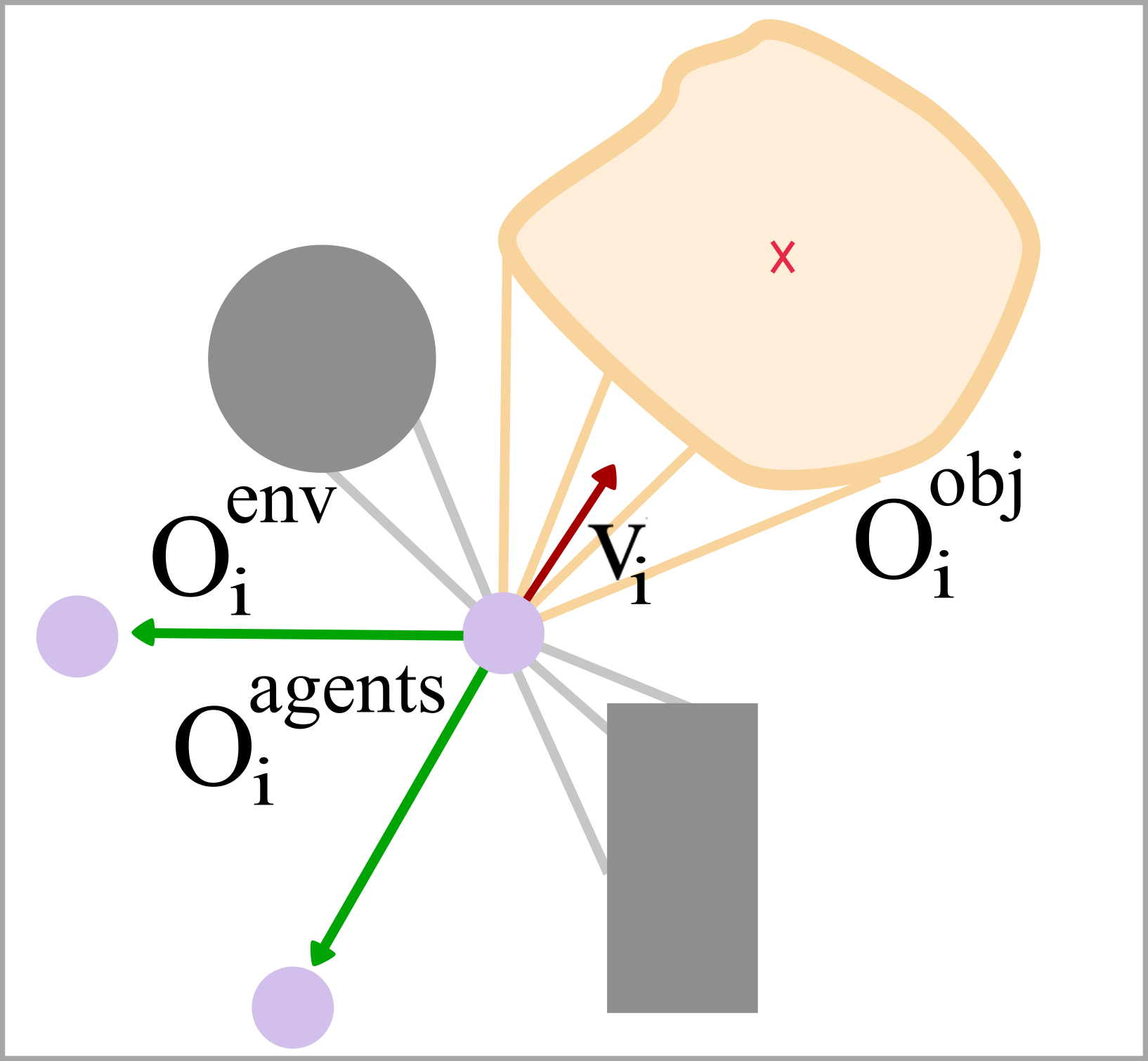}
        \label{fig:03a}
    } 
    \subfloat[Local Frame]{
        \includegraphics[width=0.49\linewidth]{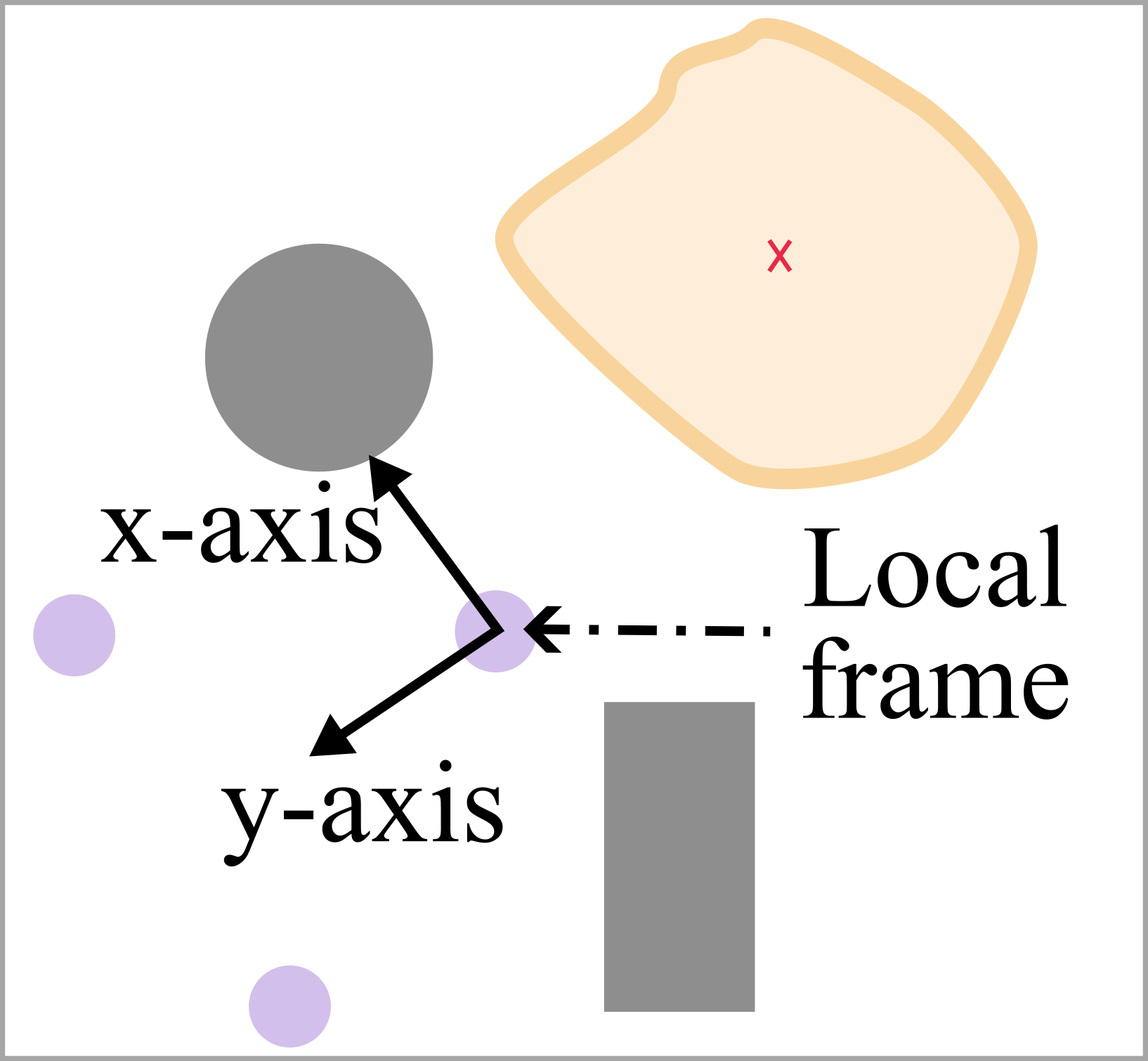}
        \label{fig:03b}
    }    
\caption{Setting description. Information about teammates \(o_i^{\text{agents}}\) , obstacles \(o_i^{\text{env}}\) and target object \(o_i^{\text{obj}}\) are green, gray, and orange lines, respectively.
Current velocity \(v_i\) is red arrow. Vectors are represented under robots’ local frames.}
    \label{fig:observation}
\end{figure}

\subsection{Problem Formulation}

We model the formation control subtask as a \emph{Partially Observable Markov Game} (POMG), which captures the decentralized nature of our multi-robot setting. 
A POMG is defined as
\begin{equation}
    \mathcal{G} = 
    \langle 
        \mathcal{N}, 
        \mathcal{S}, 
        \{\mathcal{A}_i\}, 
        P, 
        \{\mathcal{R}_i\}, 
        \{\Omega_i\}, 
        O, 
        \gamma 
    \rangle ,
\end{equation}
where
\begin{itemize}
  \item $\mathcal{N}$ is the set of agents $i = 1,\dots,N$,
  \item $\mathcal{S}$ is the global state space,
  \item $\mathcal{A}_i$ is the action space of agent $i$, and $a = (a_1, \dots, a_N)$ is the joint action,
  \item $P(s' \mid s, a)$ is the state transition function,
  \item $\mathcal{R}_i(s, a)$ is the individual reward function for agent $i$,
  \item $\Omega_i$ is the observation space of agent $i$,
  \item $O(o \mid s', a)$ is the joint observation probability function, where $o = (o_1,\dots,o_N)$, and
  \item $\gamma \in (0,1)$ is the discount factor.
\end{itemize}

At each time step~$t$, agent~$i$ receives a local observation~$o_i^t \in \Omega_i$ and selects an action~$a_i^t$ according to its policy~$\pi_{\theta_i}(a_i| o_i)$ parametrized by~$\theta_i$. The joint action~$a^t = (a_1^t,\dots,a_N^t)$ induces a state transition via~$P$, and each agent receives a reward~$r_i^t = \mathcal{R}_i(s^t, a^t)$. The objective of agent~$i$ is to maximize its expected discounted return~\(J_i(\pi)\), as defined by
\begin{equation}
J_i(\pi) = 
\mathbb{E}_{\pi}\!\left[
   \sum_{t=0}^{\infty} \gamma^{t}\, \mathcal{R}_i(s^t, a^t)
\right],
\,
\pi = (\pi_{\theta_1}, \dots, \pi_{\theta_N}).
\label{eq:agent_return}
\end{equation}

\subsection{Agents, Observations and Actions}

Agents are modeled as circular non-holonomic robots with radius \( r_a \). 
Each agent's policy~\(\pi_{\theta_i}\) outputs linear and angular velocities \( v_i\) and \(\omega_i \), which are used to update the agent's position \( (x_i, y_i) \) and heading \( \psi_i \) using unicycle kinematics. The motion model is defined by
\begin{equation}
    \dot{x}_i = v_i \cos \psi_i, \quad
    \dot{y}_i = v_i \sin \psi_i, \quad
    \dot{\psi}_i = \omega_i.
\end{equation}
Each agent receives a structured observation vector~\(o_i\), which we define as 
\begin{equation}
    o_i = [o_i^{\text{self}}, o_i^{\text{robots}}, o_i^{\text{env}}, o_i^{\text{obj}}, o_i^{S_{load}}]
\end{equation}
with self-observation~\(o_i^{\text{self}}\), observation of other robots~\(o_i^{\text{robots}}\), environment observation~\(o_i^{\text{env}}\), target object observation~\(o_i^{\text{obj}}\), and similarity score observation~\(o_i^{\text{ss}}\).

Self-observation~\(o_i^\text{self}\) is given by
\begin{equation}
o_i^{\text{self}}= \{\, p_i,\; v_i,\; \ell_i \,\}
\label{eq:self_observation}
\end{equation}
with robot position~\(p_i\), linear velocity~\(v_i\), and current load~\(\ell_i\).
The computation of a robot's current load is described in more detail in Sec.~\ref{subsec:LoadSharing}.

We define the observation of other robots~\(o_i^{\text{robots}}\) as
\begin{equation}
o_i^{\text{robots}}= \{\, \Delta p_{ij},\; \ell_j \,\}_{\, j \neq i},
\label{eq:agent_observation}
\end{equation}
where \(\Delta p_{ij} = p_j - p_i\) are the relative positions and \(\l_j\) the loads of all other robots.

Environment observation~\(o_i^{\text{env}}\) is given by
\begin{equation}
o_i^{\text{env}} = D_i^o
\label{eq:env_observation}
\end{equation}
with vector~\(D^o_i\) of \(n_r\)~equidistant distance measurements to static obstacles and agents covering \(360^\circ\) around the robot with a fixed sensing radius \(D^o_{max}\). 

The target object observation~\(o_i^{\text{obj}}\) is defined as 
\begin{equation}
o_i^{\text{obj}} = D_i^{\mathcal{P}}
\label{eq:object_observation}
\end{equation}
with \(n_r\)-dimensional vector~\(D_i^{\mathcal{P}}\) of distance measurements to the boundary of the target object's footprint~\(\mathcal{P}\).
Each distance measurement~\(d_r^{\mathcal{P}}\) is given by
\begin{equation}
d_r^{\mathcal{P}} = \min \bigl\|\, \text{ray}_r \cap \partial \mathcal{P} - p_i \,\bigr\|_2, 
\qquad r = 1,\dots,n_r
\label{eq:ray_distance}
\end{equation}
with \(\text{ray}_r\) representing the synthetic LiDAR rays. 

In addition, we include similarity score observation~$o_i^{S_{load}}$ derived from the observed load values. 
It indicates the similarity of the current robot loads to an even load distribution, as explained in detail in Sec.~\ref{subsec:LoadSharing}. 
This aggregated observation provides a compact summary of the team progress towards a successful carrying formation.

\subsection{Load Computation and Sharing} 

\label{subsec:LoadSharing}
Only robots located inside the object's footprint polygon ($\mathcal{I}=\{i \in \{1,...,n\} |\  p_i \in \mathcal{P}\} $) will participate in the object transport. For each of these robots, we compute the relative vector to the object's CoM as 
\begin{equation}
p_i^{c} = p_i - CoM = (x_i^c, y_i^c), \quad i \in \mathcal{I}.
\end{equation}
The static equilibrium conditions are given by
\begin{align}
\sum_{i\in\mathcal{I}} \ell_i = w,\quad \sum_{i\in\mathcal{I}} x_i^c \, \ell_i = 0, \quad \sum_{i\in\mathcal{I}} y_i^c \, \ell_i = 0,
\end{align}

where $w$ is the object weight. We express this system in matrix form:
\[\mathbf{A} \, \mathbf{L}_{\mathcal{I}}= B\]
\begin{equation}
\underbrace{
\begin{bmatrix}
1 & 1 & \cdots & 1 \\
x_{\mathcal{I}_1}^c & x_{\mathcal{I}_2}^c & \cdots & x_{\mathcal{I}_k}^c \\
y_{\mathcal{I}_1}^c & y_{\mathcal{I}_2}^c & \cdots & y_{\mathcal{I}_k}^c
\end{bmatrix}}_{\mathbf{A} \, (3 \times k)}
\underbrace{
\begin{bmatrix}
\ell_{\mathcal{I}_1} \\ \ell_{\mathcal{I}_2} \\ \vdots \\ \ell_{\mathcal{I}_k}
\end{bmatrix}}_{\mathbf{L}_{\mathcal{I}} \, (k \times 1)}
=
\underbrace{
\begin{bmatrix}
w \\ 0 \\ 0
\end{bmatrix}}_{\mathbf{B} \, (3 \times 1)}
\end{equation}
with the total number~\(k\) of robots inside the object footprint.

The loads vector of the robots inside the footprint \({\mathbf{L}}_{\mathcal{I}} \) is solved using least-squares for \(k \geq 3\) and the Moore-Penrose pseudoinverse otherwise, and normalized by
\begin{equation}
\mathbf{\hat{L}}_{\mathcal{I}}
= \frac{\mathbf{L}_{\mathcal{I}}}{\sum \mathbf{L}_{\mathcal{I}} + \varepsilon}\,.
\label{eq:load_normalization}
\end{equation}
This partial normalized loads vector \(\mathbf{\hat{L}}_{\mathcal{I}}\) is then padded with $n-k$ zeros to account for robots outside the footprint, which do not carry any load. This yields the full load vector 
\begin{equation}
    \mathbf{\hat{L}}=
    \begin{bmatrix}
        \mathbf{\hat{L}}_{\mathcal{I}}\\
        \mathbf{0}_{n-k}
    \end{bmatrix}.
\end{equation}

We compare this loads vector to the ideal uniform load vector \(\mathbf{u} = \frac{1}{N}{\textbf{1}_N}\).
We calculate a loads similarity score using cosine similarity, as defined by
\begin{align}
S_{load} &= \alpha \cdot \left(\frac{\boldsymbol{\hat{L}}^T \mathbf{u}}{\|\boldsymbol{\hat{L}}\|_2 \cdot \|\mathbf{u}\|_2 + \varepsilon}\right)^\beta,
\label{eq:similary_score_}
\end{align}
which can be simplified as follows 
\begin{align}
\because \boldsymbol{\hat{L}}^T \mathbf{u} &= \sum_i  \frac{1}{n} \boldsymbol{\hat{L}}_i = \frac{1}{n}, \quad \\
\& \quad \|\mathbf{u}\|_2 &= \sqrt{n \left(\frac{1}{n}\right)^2} = \frac{1}{\sqrt{n}}, \\
    \therefore S_{load} &= \alpha \left(\frac{1}{\sqrt{n} \ \|\boldsymbol{\hat{L}}\|_2}\right)^\beta.
\label{eq:similary_score}
\end{align}
Here, $\alpha > 0$ is a scaling coefficient and exponent $\beta \geq 1$ controls how sharp the reward becomes.
Larger \(\beta\)~values put more emphasis on reaching a perfectly balanced load distribution. 
The constant~$\varepsilon$ ensures numerical stability in cases where the vectors are close to zero.

\subsection{Reward Design}

We define the reward function~\(R_i\), that is calculated for each robot~\(i\), as 
\begin{equation}
R_i = R_{i}^{\text{pos}}
    + R_{i}^{\text{collide}}
    + R_{i}^{\text{loadShare}}
    + R_{i}^{\text{success}}.
\label{eq:reward_function}
\end{equation}
\(R_i^\text{pos}\) rewards robots for moving toward and getting inside the object footprint, and is defined by
\begin{equation}
    R^\text{pos} = \begin{cases} R_{in} & \text{if } p \in \mathcal{P}, \\ \alpha_{p} \cdot(d_{\mathcal{P}}^t - d_{\mathcal{P}}^{t-1}) & \text{otherwise}. \end{cases} 
\end{equation}
where $d_{\mathcal{P}}$ is the closest distance from the robot to the footprint polygon and $\alpha_{p}$ is a reward shaping factor. 

\(R^{\text{collide}}\) is a collision penalty.

\(R^\text{loadShare}\) rewards sharing the load, which is represented by the similarity score (Eq.~\ref{eq:similary_score}). 

In addition, we give the MRS a constant reward~\(R_s\) when all robots are inside the object footprint area and the similarity score exceeds the threshold $\tau_{s}$.
The success reward \(R_i^\text{success}\) is given by 
\begin{equation}
    R^\text{success} =\begin{cases}{R_{s}} & \sum_i^n 1_{p_i \in \mathcal{P}} =n \;\&\; S_{load}\geq \tau_{s},\\ 0 & \text{otherwise}. \end{cases}
\end{equation}

\begin{table}[t]
\caption{Environment Parameters. Spatial quantities are expressed in simulator units.}
\centering
\begin{tabular}{|l|l|}
\hline
\textbf{Parameter} & \textbf{Value}   \\
\hline 
\multicolumn{2}{|c|}{\textbf{Agents, Observations and Actions}} \\
\hline
$r_a$ \((N \in \{3,5\})\) & 0.05 \\
$r_a$ \((N \in \{10,15,20\})\) & 0.02 \\
$D^o_{max}$ & 0.35 \\
$n_r$ & 12 \\
$D^{\mathcal{P}}_{max}$ & 1.4 \\
$r_o$ & 0.05\\
\hline 
\multicolumn{2}{|c|}{\textbf{Load Computation and Sharing}} \\
\hline
$\alpha$ & 2 \\
$\beta$ & 10 \\
\hline 
\multicolumn{2}{|c|}{\textbf{Reward Design}} \\
\hline
$R_{in}$ & 0.1 \\
$\alpha_p$ & 0.1 \\
$R^{collide}$ & -0.5 \\
$R_s$ & 5 \\
\hline
\multicolumn{2}{|c|}{\textbf{Similarity Score Threshold}} \\
\hline
$\tau_s^3, \tau_s^5, \tau_s^{10}, \tau_s^{15}, \tau_s^{20}$ & [1.6049, 1.2999, 0.7954, 0.5086, 0.3374]\\
\hline
\multicolumn{2}{|c|}{\textbf{Obstacles}}\\
\hline
$n_{O}$ & 10 \\
$r_{O}$ & 0.05 \\
$r_{O}^{c} $ & \text{Uniform}(0.06, 0.16)\\
$w_{O}^{c} $ & \text{Uniform}(0.12, 0.35)\\
$h_{O}^{c} $ & \text{Uniform}(0.06, 0.20)\\
\hline
\multicolumn{2}{|c|}{\textbf{Objects}} \\
\hline
$\mathcal{P}_s$ & 0.4 \\
$\mathcal{P}_v$ & $\text{Uniform}_{\text{int}}(5,25)$ \\ 
$\mathcal{P}_c$ & 0.2 \\ 
$n_{cav}$ & $\text{Uniform}_{\text{int}}(1, 2)$\\ 
$k_{cav} $ & $\text{Uniform}(0.1, 0.25)$\\
\hline
\end{tabular}
\label{tab:parameters}
\end{table}

\begin{table}[t]
\caption{MAPPO Hyperparameters}
\label{tab:hyperparams}
\begin{tabular}{l|l|l}
\textbf{Parameter} & \textbf{Search Space} & \textbf{Chosen Value} \\
\hline 
$\eta$ & LogUniform($1e-5,\,3e-4$) & $9e-5$ \\
$\epsilon$ & Uniform($0.05,\,0.3$) & $0.2$ \\
$\beta_{ent}$ & LogUniform($1e-4,\,3e-2$) & $4e-3$ \\
$\lambda$ & Uniform($0.85,\,0.97$) & $0.935$ \\
$\gamma$ & $\{0.95, 0.97, 0.98, 0.99\}$ & $0.98$ \\
$K$ & $\{3, 5, 8\}$ & $8$ \\
$M$ & $\{1000, 2000, 3000, 6000\}$ & $1000$ \\
$\|\nabla\|_{\max}$ & $\{0.5, 1.0\}$ & $1.0$ \\
$L_\pi$ & $\{2, 3\}$ & $3$ \\
$L_V$ & $\{2, 3\}$ & $2$ \\
$H_\pi$, $H_V$ & $\{256, 512, 1024\}$ & $512$ \((N \in \{3, 5, 10\}\)), \\
        &                      & $1024$ \((N \in \{15, 20\})\) \\
\end{tabular}
\end{table}

\subsection{Similarity Score Threshold Choice}
We choose $\tau_{s}$ as the minimum acceptable similarity score for each team size~$N$ and desired maximum difference in carried load between agents defined as
\begin{equation}
    d = \max(\boldsymbol{\hat{L}}) - \min(\boldsymbol{\hat{L}}), \quad (0\leq d \leq 2).
\end{equation}
From (Eq.~\ref{eq:similary_score}), minimizing $S_{load}$ means maximizing $\|\boldsymbol{\hat{L}}\|_2$. There are three scenarios:
\begin{itemize}
    \item $d=0$: This is the optimal case where $\boldsymbol{\hat{L}} = \mathbf{u}$ and $\|\boldsymbol{\hat{L}}\|_2
    = \frac{1}{\sqrt{n}}$. 
    \item $d=2$: Here, the worst case will be having as many normalized \(-1\) and \(1\) load values as possible so that the total load will be \(1\). Theoretically, a load of \(-1\) would require a robot to pull down the object, which is outside of the robot's capabilities. 
    Hence, the maximum possible $\|\boldsymbol{\hat{L}}\|_2$ is:
    \begin{equation}
        \|\boldsymbol{\hat{L}}\|_{2, max}  = \begin{cases}
            \sqrt{N} & (N~\text{mod}~2 = 1), \\ \sqrt{N-1} & (N~\text{mod}~2 =0).
        \end{cases}
    \end{equation}
    \item $0 < d < 2$: The worst case will be having one agent carrying the maximum load $a$ and all the other agents carrying the minimum load $a-d$. 
    \begin{equation}
        a + (n-1) (a-d) = 1 , \quad
      \therefore a = \frac{1+d(n-1)}{n}.
    \end{equation}
    Hence, the maximum possible $\|\boldsymbol{\hat{L}}\|_2$ is:
    \begin{align}
        \|\boldsymbol{\hat{L}}\|_{2, max} = \sqrt{a^2+(n-1)(a-d)^2} \\
        = \sqrt{\frac{(1+(n-1)d)^2 + (n-1)(1-d)^2}{n^2}}.
    \end{align}
\end{itemize}
 We substitute this value back in (Eq.~\ref{eq:similary_score}) and set the result $S_{load}$ as our threshold $\tau_{s}$.

\subsection{Policy Optimization}

We use Multi-Agent Proximal Policy Optimization (MAPPO)~\cite{yu2022surprising} to optimize robot policies. 
MAPPO extends Proximal Policy Optimization (PPO)~\cite{schulman2017proximalpolicyoptimizationalgorithms} with decentralized policies (actors) and a centralized value function (critic). 
Each robot runs an identical stochastic policy that conditions only on its own observations, while the critic conditions on the joint observations of all agents following the centralized training decentralized execution paradigm (see Fig.~\ref{fig:algorithm}).

We parameterize the actor with a shared-parameter multi-agent multi-layer perceptron (MLP) that takes each robot's observation vector as input and outputs the parameters of a diagonal Gaussian policy. 
The network has three hidden layers and ReLU activations. 
The Gaussian action is squashed through a $\tanh$ function to respect the action bounds of the environment. 
All robots share the same actor parameters, reflecting their homogeneous nature.

The critic is implemented as a multi-agent MLP with two hidden layers and ReLU activations. 
It is centralized, that is, it receives the concatenated observations of all robots as input and outputs a separate state-value estimate for each robot. 
This centralized critic is used to compute generalized advantage estimates for PPO.

\section{EXPERIMENTAL SETUP}

\subsection{Environment}

\begin{figure*}[t]
    \centering
    \subfloat[E1]{
        \fbox{\includegraphics[width=0.18\textwidth, trim=20 20 20 20, clip]{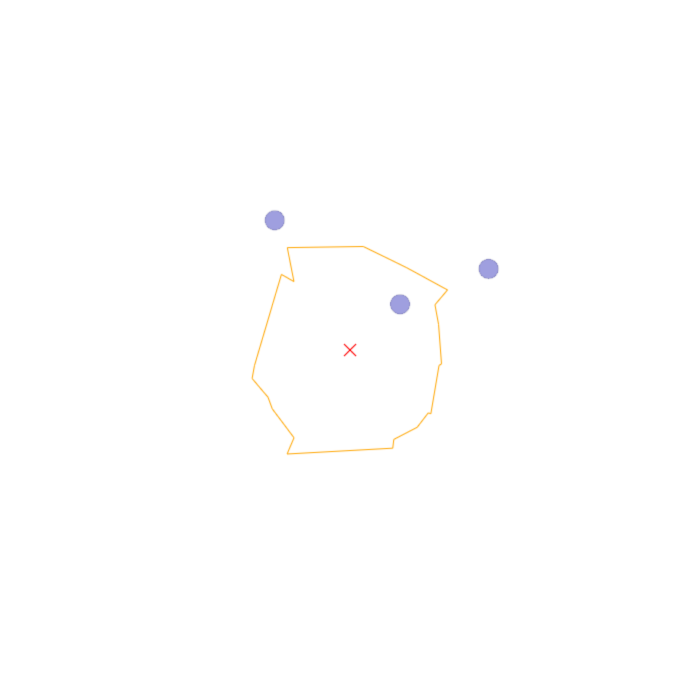}}
        \label{fig:04a}
    }   
    \subfloat[E2]{
        \fbox{\includegraphics[width=0.18\textwidth, trim=20 20 20 20, clip]{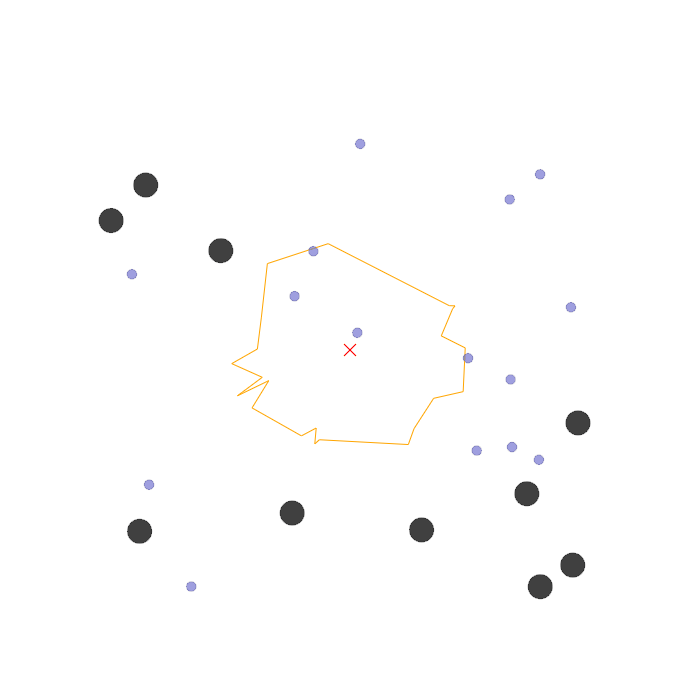}}
        \label{fig:04b}
    }    
    \subfloat[E3]{
        \fbox{\includegraphics[width=0.18\linewidth, trim=20 20 20 20, clip]{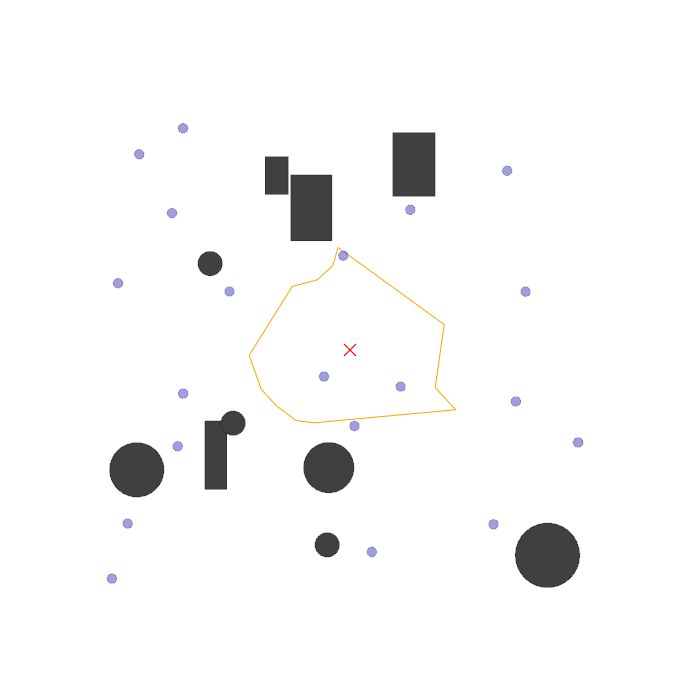}}
        \label{fig:04c}
        }    
    \subfloat[E4]{
        \fbox{\includegraphics[width=0.18\linewidth, trim=20 20 20 20, clip]{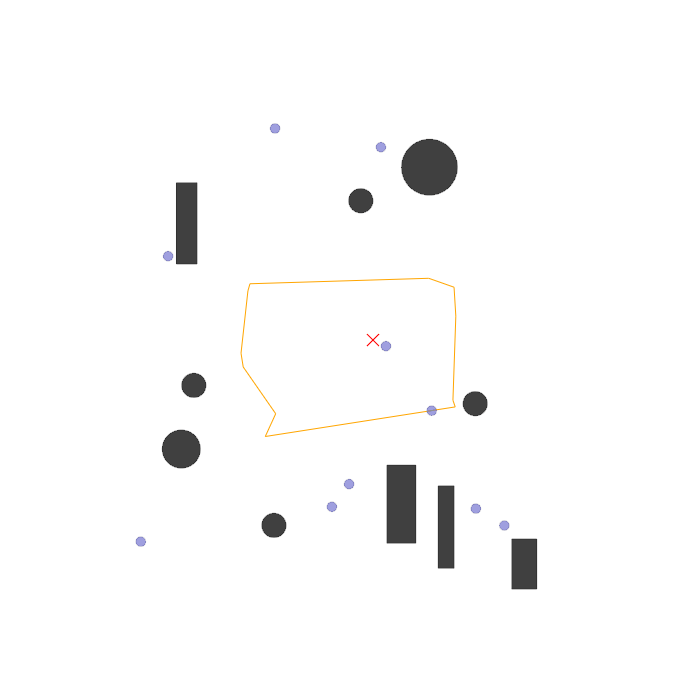}}
        \label{fig:04d}
        }
    \subfloat[E5]{
        \fbox{\includegraphics[width=0.18\linewidth, trim=20 20 20 20, clip]{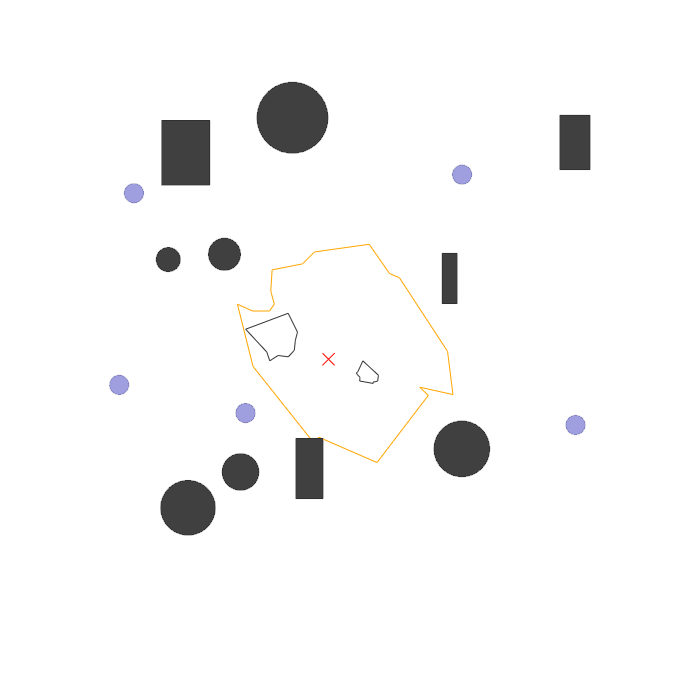}}
        \label{fig:04e}
    }
    \caption{Environments E1 - E5 (purple circles: robots, red cross: CoM, orange lines: object footprint, black circles and rectangles: obstacles)}
    \label{fig:envs}
\end{figure*}

We conduct our experiments in the 2D simulator VMAS~\cite{bettini2024vmas} and evaluate our approach across five environment variants of increasing difficulty, see Fig.~\ref{fig:envs}. 
The first environment (E1) is an obstacle-free scene with a simple object with uniform mass distribution (Fig.~\ref{fig:04a}). The object is generated as a random star-shaped polygon parameterized by its size ($\mathcal{P}_s$), number of vertices ($\mathcal{P}_{v}$) and concavity ($\mathcal{P}_c \in [0,1]$). The concavity parameter controls the amplitude of radial perturbations used to construct the polygon. When $\mathcal{P}_c = 0$, the object reduces to a regular convex polygon. 
As \(\mathcal{P}_c\) increases, larger radial variations are introduced, increasing the likelihood of concave geometries.
The radius~\(r_i\) of each vertex~\(i\) is sampled as:
\begin{equation}
r_i = 1 + 0.9 \  \mathcal{P}_c \  \mathcal{U}(-1,1)
\end{equation}
The factor of~\(0.9\) prevents zero radii while still allowing substantial geometric variation.
The second environment (E2) extends this setup by adding $n_O$ circular obstacles of radius $r_O$ at random positions in the scene, requiring the robots to avoid basic clutter during the pattern formation task (Fig.~\ref{fig:04b}). 
In the third environment (E3), we further increase the complexity by introducing a denser mixture of circular and rectangular obstacles with varying sizes, creating tighter passages and more irregular free space, while still keeping the simple object (Fig.~\ref{fig:04c}). Circular obstacles are parameterized by their radius $r_O^c$, whereas rectangular obstacles are defined by their width $w_O^c$ and height $h_O^c$.
In environment four (E4), we simulate a more complex object with non-homogeneous mass distribution by randomly shifting the object's CoM (Fig.~\ref{fig:04d}). This shift is sampled from a normal distribution with $\sigma=0.1*(\mathcal{P}_s/2)$. 
The last environment (E5) uses the complex object from E4 and additionally adds $n_{cav}$~cavities to the object (Fig.~\ref{fig:04e}).
The cavities are generated similarly to the main polygon with a scaling factor~$k_{cav}$  and randomly placed within the main object polygon.
Table~\ref{tab:parameters} gives the environment parameters.

\subsection{Hyperparameters}

For each MRS size $N \in \{3, 5, 10, 15, 20\}$, we train separate policies to study how our approach scales with the number of robots.
We train using \num{150}~parallel random environments, each simulated for \num{200}~steps per episode, resulting in batches of \num{30000}~environment steps per policy update. We perform \num{125}~policy updates, for a total of $3.75 \times 10^{6}$~environment steps. 
For $N\in\{15,20\}$, we use \num{300}~steps per episode to give the robots enough time to find a successful formation.
While training, we create random environments with similar properties as E1 - E5 by randomly placing up to $n_o$~obstacles as in E3, randomly shifting the CoM as in E4, and introducing objects with cavities as in E5 with probability $\text{P}_{cav} = 0.35$.
We run the training with a fixed random seed (\texttt{seed = 0}) for reproducibility.

We conduct a randomized hyperparameter search using Ray Tune~\cite{liaw2018tune} for setting MAPPO's hyperparameters. 
In particular, we select clipping parameter~$\epsilon$, discount factor~$\gamma$, and GAE parameter~$\lambda$ for the clipped surrogate objective function; entropy coefficient~$\beta_{ent}$, epoch reuse~\(K\) per collected batch, and mini-batch size~\(M\) for sampling from the replay buffer.
We normalize advantages per robot. 
We optimize actor and critic parameters jointly using Adam and select learning rate~$\eta$ and apply gradient clipping with a maximum~$\ell_2$ norm, $||\nabla||_{\max}$. 
For each \(N\), we tested a total of 64 independent configurations trained for \(50\)~iterations each and selected the final configuration based on the mean episode reward.
The search spaces and chosen values for each hyperparameter are shown in Table \ref{tab:hyperparams}.

\subsection{Post-Evaluation}
\label{subsec:posteval}


We post-evaluate the trained policies across environments E1 - E5 in \(150\)~runs each with maximum 400 steps per run. 
We report two metrics:
The Success Rate (SR) gives the percentage of runs in which the maximum normalized load difference between robots remains below 0.15, that is, when the load similarity exceeds the predefined threshold and the MRS receives the success reward. 
The Mean Execution Steps (MES) metric gives the mean number of environment steps required to reach this successful formation.

In addition, we compare the resulting robot formations of our approach against two baselines for pattern formation: Dynamic Uniform Distribution (DUD) + Expectation Maximization (EM) and a randomized formation. 
In Dynamic Uniform Distribution (DUD) + Expectation Maximization (EM)~\cite{Bi2018SOSF}, robots are first driven into the target shape using an attraction field derived from the gradient of a distance transform image, while mutual repulsive forces promote an approximately uniform distribution inside the shape (DUD). This dynamic configuration serves as an initialization for an EM-like refinement stage, in which each agent iteratively moves to the centroid of its assigned region, improving uniformity.
As a neutral baseline, we include a randomized formation generated within the footprint by sampling $x$ and $y$ coordinates independently from uniform distributions.

To quantify performance in the baseline comparisons, we compute the $\mathtt{L_1}$ transportation cost required to move from the final formation  $\mathtt{P}=\{p_{i}\}^N_{i=1}$  to the closest optimal formation $\mathtt{P^{*}}=\{p^{*}_{i}\}^N_{i=1}$, where all robots carry equal load and satisfy a minimum separation constraint. 
To compute this closest optimal formation, we restrict the footprint to randomly created simpler shapes represented as the union of 2–5 overlapping rectangles. This representation allows us to formulate a mixed integer linear programming (MILP) problem that determines the closest optimal target configuration $\mathtt{P^*}$. The MILP minimizes the total $\mathtt{L_1}$ norm (Manhattan distance) between the current robot positions and the optimal ones while enforcing equal load distribution and minimum inter-robot spacing.

We do \(100\)~runs in an obstacle-free environment and objects with randomly shifted CoM per method and MRS size \(N \in \{3,5,10,15,20\}\).
Please note that we scaled the object by a factor of 2 for MRS sizes~$N\in\{15,20\}$.

\section{RESULTS}

\begin{figure}[t]
    \centering\includegraphics[width=0.85\linewidth]{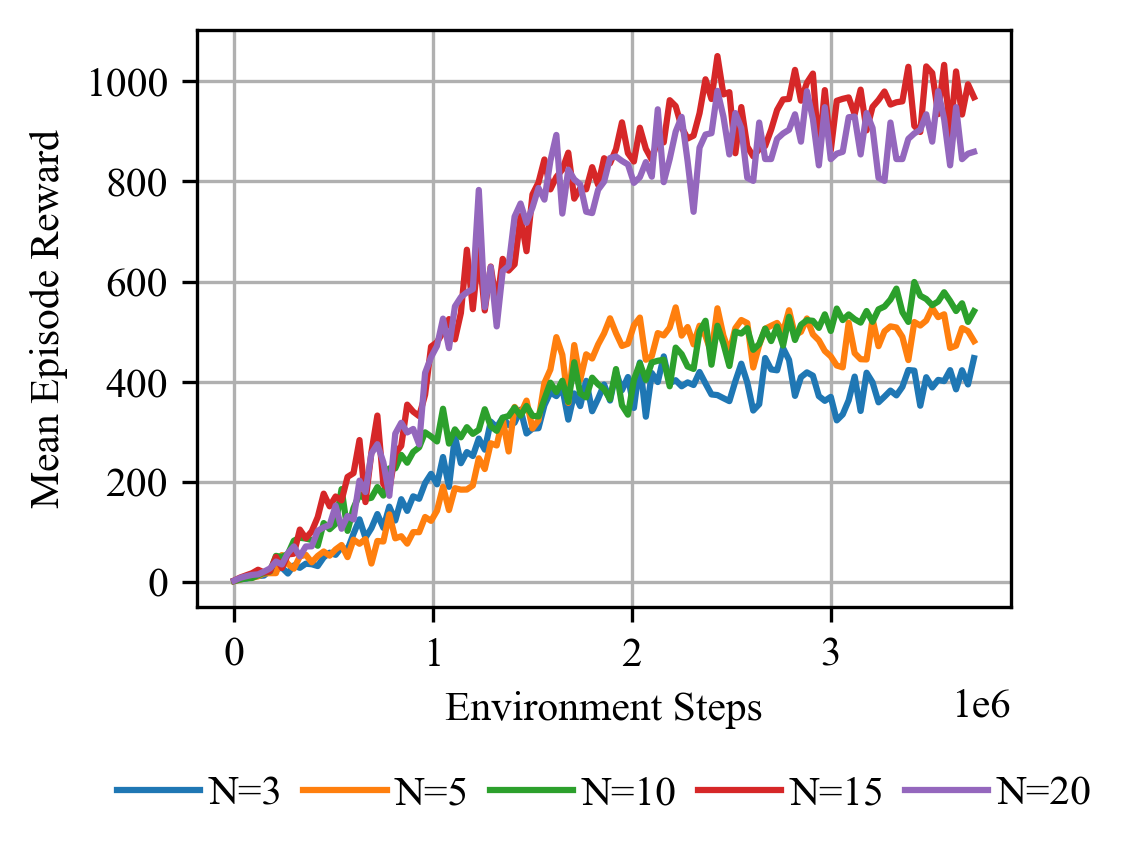}
    \label{fig:06}
    \caption{Learning Curves}
    \label{fig:learningcurve}
\end{figure}

\subsection{Policy Optimization}

Figure~\ref{fig:learningcurve} illustrates the learning curves of the trained policies for different MRS sizes~\(N\). All configurations show stable performance improvement, indicating effective policy optimization. For \(N=3\), the mean episode reward remains lower compared to~\(N\in\{5, 10\}\). This behavior is primarily due to the load-sharing reward component (Eq. 19), which is computed as a high-order cosine similarity between the actual and ideal load distributions. In small MRS sizes, torque balance around the center of mass is geometrically constrained, and each robot’s contribution significantly affects stability. Consequently, achieving near-uniform load distribution is more restrictive, making successful episodes less frequent and limiting cumulative reward.
In contrast, larger MRS sizes~\(N\in\{15, 20\}\) benefit from redundancy in load distribution, which relaxes geometric constraints and facilitates faster convergence. 
It is important to note that policies for these two MRS sizes were trained with a maximum episode length of 300 steps, whereas smaller MRS sizes used 200 steps. The extended episode length enables greater accumulation of positive shaping rewards \(R^\text{pos}\) and \(R^\text{success}\), resulting in higher absolute returns. Therefore, raw episode rewards are not directly comparable across different MRS sizes. Instead, policy quality is evaluated through post-evaluation metrics such as success rate, execution steps, and formation cost (see Tables III–IV).
Overall, the learning curves demonstrate stable convergence across all MRS sizes and confirm that the proposed training formulation scales effectively with increasing number of robots.

\subsection{Post-Evaluation}

\begin{table*}[t]
\caption{Success Rate (SR) and Mean Execution Steps (MES) for different MRS sizes}
\centering
\renewcommand{\arraystretch}{1.5}
\begin{tabular}{|c|cc|cc|cc|cc|cc|}
\hline
\multirow{3}{*}{\textbf{Environment}} 
& \multicolumn{10}{c|}{\textbf{No. of Robots}} \\
\cline{2-11}
& \multicolumn{2}{c|}{\textbf{3}} 
& \multicolumn{2}{c|}{\textbf{5}} 
& \multicolumn{2}{c|}{\textbf{10}} 
& \multicolumn{2}{c|}{\textbf{15}}
& \multicolumn{2}{c|}{\textbf{20}} \\
& SR & MES & SR & MES & SR & MES & SR & MES & SR & MES \\
\hline 
E1  & 94.00\% & 62.6 & 100\% & 59.23  & 98.67\% & 62.69 & 99.33\% & 47.23 & 99.33\% & 49.54 \\
E2  & 93.07\% & 62.8 & 96.67\% & 66.38  & 96.00\% & 63.62 & 99.8\% & 47.63 & 99.33\% & 50.09 \\
E3  & 92.33\% & 66.17 & 98.68\% & 66.26  & 94.00\% & 63.43 & 96.67\% & 49.52 & 95.33\% & 58.03 \\
E4 & 92.00\%  & 85.78 & 98.67\% & 80.83    & 93.33\% & 83.11 & 94.67\% & 70.71 & 94.60\% & 70.37 \\
E5 & 73.67\%  & 82.48 & 84.67\% & 105.10 & 86.67\% & 94.62 & 93.33\% & 74.27 & 91.33\% & 77.31 \\
\hline
\end{tabular}
\label{tab:final_results}
\end{table*}

\begin{table*}
\caption{Comparison with baselines across different MRS sizes}
\centering
\renewcommand{\arraystretch}{1.4}
\begin{tabular}{|c|c|c|c|c|c|c|}
\hline
\multirow{2}{*}{\textbf{Metric}} 
& \multirow{2}{*}{\textbf{Method}} 
& \multicolumn{5}{c|}{\textbf{No. of Robots}} \\
\cline{3-7}
& & 3 & 5 & 10 & 15 & 20 \\
\hline
\multirow{3}{*}{Avg. $\mathtt{L_1}$ Cost}
& Random & 0.56$\pm$0.29 & 0.90$\pm$0.45  & 1.49$\pm$0.81 & 3.89$\pm$1.73 & 4.53$\pm$2.37 \\
& DuD+EM & 0.25$\pm$0.25 & 0.40$\pm$0.23 & 0.75$\pm$0.41 & 2.48$\pm$1.11 & 3.23$\pm$1.66 \\
& Ours & \textbf{0.03$\pm$0.03} & \textbf{0.07$\pm$0.08} & \textbf{0.42$\pm$0.29} & \textbf{0.84$\pm$0.42} & \textbf{1.11$\pm$0.78} \\
\hline
\multirow{3}{*}{Success Rate}
& Random & 5.00\% & 9.00\% & 33.00\% & 49.00\% & 77.00\% \\
& DuD+EM & 14.00\% & 49.00\% & 79.00\% & 91.00\% & 96.00\% \\
& Ours & \textbf{90.00\%} & \textbf{83.00\%} & \textbf{93.00\%} & \textbf{95.00\%} & \textbf{97.00\%} \\
\hline
\multirow{3}{*}{Avg. $S_{load}$}
& Random & 0.32$\pm$0.55 & 0.39$\pm$0.54 & 0.63$\pm$0.62 & 0.72$\pm$0.59 & 0.93$\pm$0.58 \\
& DuD+EM & 1.03$\pm$0.54 & 1.19$\pm$0.56 & 1.27$\pm$0.51 & 1.23$\pm$0.47 & 1.30$\pm$0.46 \\
& Ours & \textbf{1.84$\pm$0.41} & \textbf{1.63$\pm$0.63} & \textbf{1.55$\pm$0.44} & \textbf{1.28$\pm$0.44} & \textbf{1.27$\pm$0.56} \\
\hline
\end{tabular}
\label{tab:comparison}
\end{table*}

During post-evaluation (see Sec.~\ref{subsec:posteval}), all MRS sizes achieve high success rates across the first four environments, indicating that the learned policies generalize well to variations in obstacle configurations and non-homogeneous mass distributions, see Tab.~\ref{tab:final_results}.
However, performance drops for E5, where the presence of cavities makes the object geometry more irregular and reduces the feasible regions for robot placement, which in turn increases the difficulty of achieving balanced load distribution. As before, this effect is more pronounced for \(N=3\)~robots.

The MES values provide additional insight into transport efficiency. For small MRS sizes~\(N\in\{3, 5, 10\}\), the required number of execution steps is relatively higher than for larger MRS sizes, particularly in E4 and E5, reflecting the increased effort needed to achieve stable load balance. In contrast, MES decreases substantially across all environments for \(N\in\{15, 20\}\). This indicates that larger MRS sizes converge more quickly to stable transport formations, reducing corrective repositioning and oscillatory behavior.

Table IV compares our method with the two baselines. While Random initialization performs poorly across all metrics, DuD+EM achieves moderate improvements due to its geometrically uniform distribution strategy. However, DuD+EM does not account for non-uniform mass distributions or torque balance constraints. As a result, the average \(\mathtt{L_1}\) formation cost grows significantly with increasing MRS size for DuD+EM, whereas our method maintains substantially lower deviation from the MILP-optimal solution. We also find that success rates for our method remain consistently high across all MRS sizes, particularly in small MRS settings, where geometry-only strategies struggle. Our approach also achieves higher average load similarity score ($S_{load}$), especially for small MRS sizes, indicating more effective load distribution.

\begin{figure*}[t]
    \centering
    \includegraphics[width=0.65\textwidth]{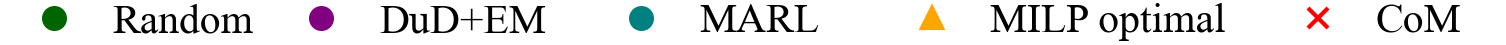}
    \subfloat[Random]{
        \includegraphics[width=0.32\textwidth]{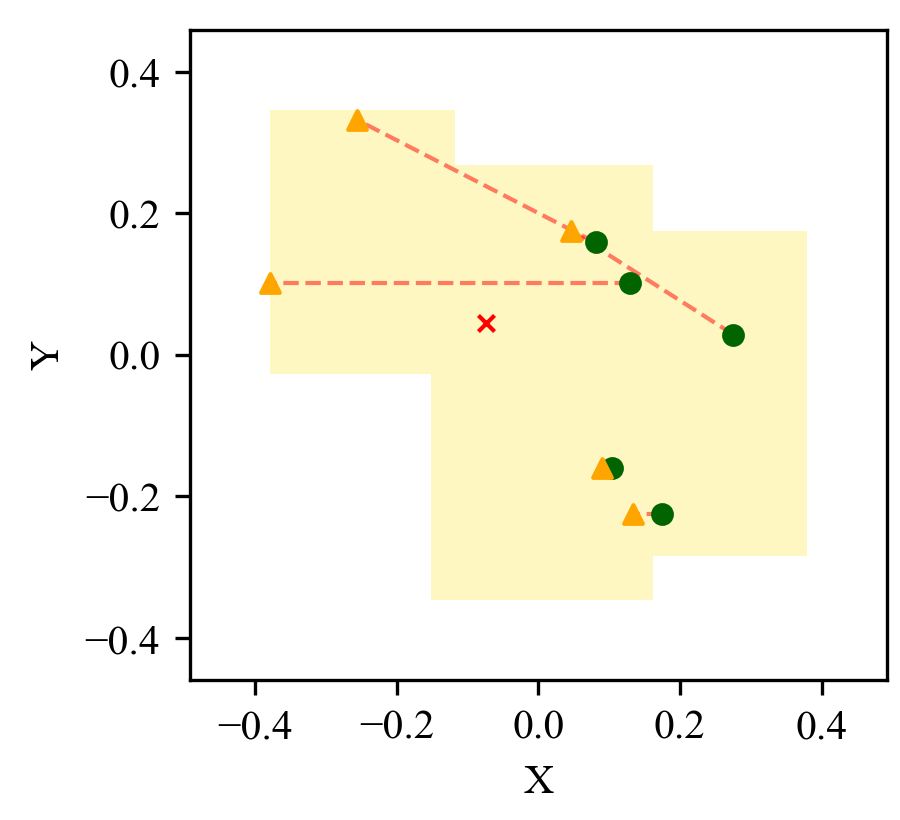}
        \label{fig:comp_random}
    }   
    \subfloat[DUD+EM]{
        \includegraphics[width=0.32\textwidth]{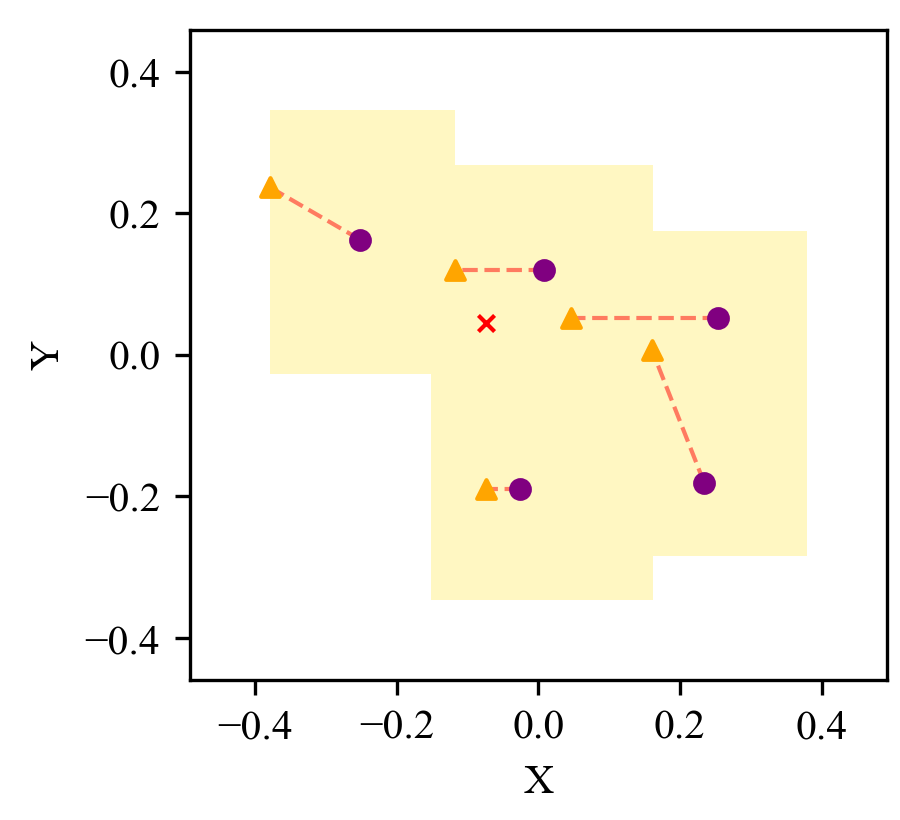}
        \label{fig:comp_dudem}
    }    
    \subfloat[Our method]{
        \includegraphics[width=0.32\linewidth]{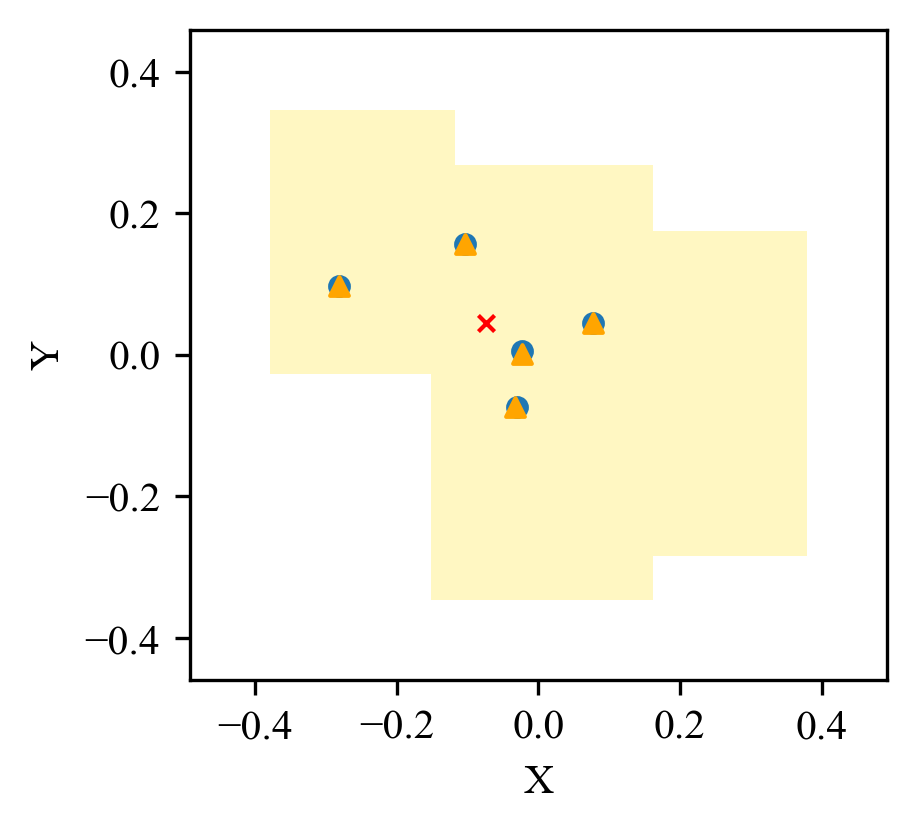}
        \label{fig:comp_marl}
    }   
    \caption{Visualization of the transportation cost to the closest optimal formation for \(N=5\)~robots carrying an object with non-uniform mass distribution. }
    \label{fig:comparison}
\end{figure*}

Finally, Figure~\ref{fig:comparison} provides a qualitative comparison for \(N=10\) robots transporting an arbitrary shaped object with non-uniform mass distribution. The pattern by the Random method shows scattered placements and poor alignment with the MILP-optimal solution. DuD+EM produces a geometrically uniform distribution. 
However, it fails to properly compensate for asymmetric mass distribution.
In contrast, our learned policy closely matches the MILP-optimal formation. 

Overall, we find that our policies implicitly learn to approximate the combinatorial optimization solution through decentralized control.

\section{CONCLUSION}

In this paper, we present a framework based on multi-agent reinforcement learning to load-aware cooperative object transport. We specifically addressed the formation phase preceding transport and trained decentralized robot policies using multi-agent proximal policy optimization within a centralized training decentralized execution paradigm. The resulting policies enable robots to navigate cluttered environments, enter an object’s footprint, and self-organize into stable support configurations for objects with both regular and highly irregular geometries and non-uniform mass distributions.

Post-evaluation across multiple environments and numbers of robots demonstrates that the proposed approach achieves high success rates, small execution steps, and scalability with increasing numbers of robots. Moreover, the learned formations closely approximate mixed integer linear programming-optimal solutions, indicating that the policies implicitly capture the underlying load-balancing structures. Our results show that explicitly incorporating load-sharing objectives into the learning formulation enables coordinated multi-robot transport beyond purely geometric strategies.
To the best of our knowledge, our work is the first to explicitly address non-uniform object mass distribution during pattern formation in object-carrying tasks, while also ensuring the participation of all robots in transportation.

Despite our encouraging results, there as several aspects that warrant future research. We aim at extending the framework toward full transport execution after lifting, integrating onboard estimation of object geometry and mass distribution, and investigating more scalable policy architectures such as graph-based representations to further improve coordination in large-scale multi-robot systems.

\section*{ACKNOWLEDGEMENTS}

This work has been partially supported by the German Federal Ministry of Research, Technology and Space (BMFTR) under the Robotics Institute Germany (RIG).
The authors gratefully acknowledge the scientific support and HPC resources provided by the Erlangen National High Performance Computing Center (NHR@FAU) of the Friedrich-Alexander-Universität Erlangen-Nürnberg (FAU) under the BayernKI project v106be BayernKI funding is provided by Bavarian state authorities.


\bibliographystyle{IEEEtran}
\bibliography{references.bib}

\end{document}